\documentclass{article}

\usepackage{amsmath, amssymb}

\usepackage{microtype}
\usepackage{graphicx}
\usepackage{subfigure}
\usepackage{booktabs} 

\usepackage{hyperref}

\usepackage[accepted]{sysml2019}  

\sysmltitlerunning{Accelerated Methods for Deep Reinforcement Learning}

\begin{document}

\twocolumn[
\sysmltitle{Accelerated Methods for Deep Reinforcement Learning}



\sysmlsetsymbol{equal}{*}

\begin{sysmlauthorlist}
\sysmlauthor{Adam Stooke}{berk}
\sysmlauthor{Pieter Abbeel}{berk}
\end{sysmlauthorlist}

\sysmlaffiliation{berk}{Department of Computer Science, University of Berkeley, Berkeley, California}

\sysmlcorrespondingauthor{Adam Stooke}{adam.stooke@berkeley.edu}

\sysmlkeywords{Deep Reinforcement Learning, Parallel Computing}

\vskip 0.3in

\begin{abstract}
Deep reinforcement learning (RL) has achieved many recent successes, yet experiment turn-around time remains a key bottleneck in research and in practice.  We investigate how to optimize existing deep RL algorithms for modern computers, specifically for a combination of CPUs and GPUs.  We confirm that both policy gradient and Q-value learning algorithms can be adapted to learn using many parallel simulator instances.  We further find it possible to train using batch sizes considerably larger than are standard, without negatively affecting sample complexity or final performance.  We leverage these facts to build a unified framework for parallelization that dramatically hastens experiments in both classes of algorithm.  All neural network computations use GPUs, accelerating both data collection and training.  Our results include using an entire DGX-1 to learn successful strategies in Atari games in mere minutes, using both synchronous and asynchronous algorithms.
\end{abstract}
]



\printAffiliationsAndNotice{} 

\section{Introduction}
\label{sec:intro}

Research in deep reinforcement learning (RL) has relied heavily on empirical evaluation, making experiment turnaround time a key limiting factor.  Despite this critical bottleneck, many reference implementations do not fulfill the potential of modern computers for throughput, unlike in supervised learning (see e.g. \cite{Imagenet_1_Hour}).  In this work, we study how to adapt deep RL algorithms--without changing their underlying formulations--to better leverage multiple CPUs and GPUs in one machine.  The result is a significant gain in efficiency and scale of hardware utilization and hence in learning speed.

Today's leading deep RL algorithms have roughly clustered into two families: (i) Policy gradient methods, of which Asynchronous Advantage Actor-Critic (A3C) \cite{Mnih2016a3c} is a representative example, (ii) Q-value learning methods, a representative example being Deep Q-Networks (DQN) \cite{mnih2015human}.  Traditionally, these two families appear in distinct implementations and use different hardware resources; in this paper we unify them under the same framework for scaling.

Our contribution is a framework for parallelized deep RL including novel techniques for GPU acceleration of both inference and training.  We demonstrate multi-GPU versions of the following algorithms: Advantage Actor-Critic \cite{Mnih2016a3c},
Proximal Policy Optimization (PPO) \cite{Schulman2017ppo}, DQN \cite{mnih2015human},
Categorical DQN \cite{Bellemare2017distributional}, and Rainbow \cite{Hessel2017rainbow}.  To provide calibrated results, we test our implementations in the heavily benchmarked Atari-2600 domain via the Arcade Learning Environment (ALE) \cite{Bellemare2012ale}.  

We found that highly parallel sampling using batched inferences can accelerate experiment turn-around time of all algorithms without hindering training.  We further found that neural networks can learn using batch sizes considerably larger than are standard, without harming sample complexity or final game score.

Beyond exploring these new learning regimes, we leverage them to dramatically speed up learning.  For example, policy gradient algorithms ran on an 8-GPU server learned successful game strategies in under 10 \textit{minutes}, rather than hours.  We similarly reduced the duration of some standard Q-value-learning runs from 10 days to under 2 hours.  Alternatively, independent RL experiments can run in parallel with high aggregate throughput per computer.  We believe that these results promise to accelerate research in deep RL, and we suggest directions for further investigation and development.

\section{Related Work}
\label{sec:related}
Efforts to parallelize and accelerate deep RL algorithms have been underway for several years.  Gorila \cite{Nair2015gorila} parallelized DQN using distributed computing.  It achieved significant although sub-linear speedups using hundreds of computing units as samplers or learners, with central parameter servers for managing parameter updates.  This effort suffered in sample complexity relative to single-threaded DQN.  More recently, \cite{horgan2018distributed} 
showed that a distributed, prioritized replay buffer can support faster learning while using hundreds of CPU cores for simulation and a single GPU for training.  The same work used increased batch sizes, with a brief study of the effect of learning rate.

The policy gradient method A3C is itself a parallelized algorithm.  In GA3C \cite{babaeizadeh2017ga3c}, a speedup over CPU-only A3C was achieved by using a GPU.  It was employed asynchronously, with ``predictor'' and ``trainer'' threads queuing observations and rewards for batched inferences and training updates.  GA3C induced a ``policy lag'' between generation and consumption of training data, worsening sample complexity.  In independent work simultaneous to ours, \cite{IMPALA} extended policy gradient methods to a distributed setting, enabling an alternative approach to multi-GPU training called IMPALA.  They introduced a more heavily modified algorithm, \textit{V-trace}, to mitigate policy lag--which we avoid--and did not employ GPU inference.  In PAAC \cite{ClementeMC17}, the authors explored the use of many simulators and increased batch sizes learning rates in (single-GPU) batched A2C--ideas central to our studies.  Our contributions to actor-critic methods exceed this work in a number of ways, chiefly: improved sampling organization, tremendously enhanced scale and speed using multiple GPUs, and inclusion of asynchronous optimization.

\section{RL Algorithm Background}
\label{sec:background}
In a standard RL formulation as a Markov Decision Process, a learning agent aims to maximize the sum of discounted rewards experienced while interacting with an environment: $R_t=\sum_{k=0}^\infty{\gamma^k r_{t+k}}$, where $r$ is the reward and $\gamma\leq 1$ the discount factor.  The value of a state, $V(s_t)=\mathbb{E}\left[R_t|s_t\right]$, is defined as the expected return under a given policy.  The Q-value, $Q(s_t,a_t)=\mathbb{E}\left[R_t|s_t,a_t\right]$ is the same but first using action $a_t$ to advance.

In policy gradient methods, the policy is directly parameterized as a distribution over actions, as $\pi(a|s;\theta)$.  The \textbf{Advantage Actor-Critic} algorithm (see, e.g. \cite{Mnih2016a3c}) learns to estimate state values $V(s;\theta)$, and iteratively optimizes the policy on fresh environment experience using gradient steps as $\mathbb{E}\left[\nabla_\theta\log\pi(a_t|s_t;\theta)A_t\right]$, where $A(s,a)=Q(s,a)-V(s)$ is the advantage, estimated as $R_t - V(s_t)$.  \textbf{Proximal Policy Optimization} (PPO) \cite{Schulman2017ppo} maximizes a surrogate objective $\mathbb{E}\left[\rho_t(\theta)A_t\right]$, where $\rho_t(\theta)=\pi(a_t|s_t;\theta)/\pi(a_t|s_t;\theta_{old})$ is the likelihood ratio of the recorded action between the updated and sampled policies.  We use the clipping-objective version of PPO, which optimizes the expression $\mathbb{E}\left[\min\left(\rho_t(\theta)A_t, clip(\rho_t(\theta),1-\epsilon,1+\epsilon)A_t\right)\right]$ under hyperparameter $\epsilon<1$.  Unlike A3C, PPO performs multiple parameter updates using (minibatches from) each set of sampled experience.

Q-value learning methods instead parameterize the Q-function $Q(s,a;\theta)$, which in \textbf{DQN} \cite{mnih2015human} is regressed against an objective as: $\mathbb{E}[\left(y_i-Q(a_i|s_i;\theta)\right)^2]$, where $y_i$ is the data-estimated Q-value given by $y_i=r_i + \gamma \max_a Q(a|s_{i+1};\theta^-)$.  The target network $\theta^-$ is periodically copied from $\theta$.  Training data is selected randomly from a replay buffer of recent experiences, each to be used multiple times. \textbf{Categorical DQN} \cite{Bellemare2017distributional} discretizes the possible Q-values into a fixed set and learns a distribution for each $Q(a|s;\theta)$.  The use of distributional learning was combined with five other enhancements under the name Rainbow: 1) Double-DQN \cite{van2016double}, 2) Dueling Networks \cite{Wang2015dueling}, 3) Prioritized Replay \cite{Schaul2016prioritized}, 4) n-step learning \cite{peng1994incremental}, and 5) NoisyNets \cite{fortunato2018noisy}.  In our experiments, we use the $\epsilon$-greedy version of Rainbow, without parameter noise: \textbf{$\epsilon$-Rainbow}.  We refer the interested reader to the original publications for further details on these algorithms.

\section{Parallel, Accelerated RL Framework}
\label{sec:framework}

We consider CPU-based simulator-environments and policies using deep neural networks.  We describe here a complete set of parallelization techniques for deep RL that achieve high throughput during both sampling and optimization.  We treat GPUs homogeneously; each performs the same sampling-learning procedure.  This strategy scales straightforwardly to various numbers of GPUs.

\subsection{Synchronized Sampling}

We begin by associating multiple CPU cores with a single GPU.  Multiple simulators run in parallel processes on the CPU cores, and these processes perform environment steps in a synchronized fashion.  At each step, all individual observations are gathered into a batch for inference, which is called on the GPU after the last observation is submitted.  The simulators step again once the actions are returned, and so on, as in \cite{ClementeMC17}.  System shared memory arrays provide fast communication between the action-server and simulator processes.

Synchronized sampling may suffer slowdowns due to the straggler effect--waiting for the slowest process at each step.  Variance in stepping time arises from varied computation loads of different simulator states and other random fluctuations.  The straggler effect worsens with increased number of parallel processes, but we mitigate it by stacking multiple, independent simulator instances per process.  Each process steps all its simulators (sequentially) for every inference batch.  This arrangement also permits the batch size for inference to increase beyond the number of processes (i.e. CPU cores).  A schematic is shown in Figure~\ref{fig:sampler_diagram}.  Slowdowns caused by long environment resets can be avoided by resetting only during optimization pauses.\footnote{For example, one may either 1) ignore a simulator in need of reset or 2) immediately swap in a fresh instance held in reserve.}

If simulation and inference loads are balanced, each component will sit idle half of the time, so we form two alternating groups of simulator processes.  While one group awaits its next action, the other steps, and the GPU alternates between servicing each group.  Alternation keeps utilization high and furthermore hides the execution time of whichever computation is the quicker of the two.

We organize multiple GPUs by repeating the template, allocating available CPU cores evenly.  We found it beneficial to fix the CPU assignment of each simulator process, with one core reserved to run each GPU.  The experiments section contains measurements of sampling speed, which increases with the number of environment instances.

\subsection{Synchronous Multi-GPU Optimization}
In our synchronous algorithms, all GPUs maintain identical parameter values.  We leverage the data-parallelism of stochastic gradient estimation and use the well-known update procedure, on every GPU: 1) compute a gradient using locally-collected samples, 2) all-reduce the gradient across GPUs, 3) use the combined gradient to update local parameters.  We use the NVIDIA Collective Communication Library for fast communication among GPUs.    

\subsection{Asynchronous Multi-GPU Optimization}
In asynchronous optimization, each GPU acts as its own sampler-learner unit and applies updates to a central parameter store held in CPU memory.  Use of accelerators compels a choice of where to perform the parameter update.  In our experience, applying common update rules to the network is faster on the GPU.  Our general update procedure includes three steps: 1) compute the gradient locally and store it on the GPU, 2) pull current central parameters onto the GPU and apply the update rule to them using the pre-computed gradient, 3) write the updated parameters back to the central CPU store. 
After this sequence, the local GPU parameters are in sync with the central values, and sampling proceeds again.  Following \cite{Mnih2016a3c}, we also centralize the update rule parameters.

Rather than add update increments to the central parameters, which requires CPU computation, we overwrite the values.  Therefore, we employ a lock around steps (2) and (3) above, preventing other processes from reading or writing parameter values concurrently.  We divide the parameters into a small number of disjoint chunks which are updated separately, each with its own lock (steps 2-3 become a loop over chunks).  This balances update call efficiency against lock contention and can provide good performance.\footnote{e.g., for 8 workers and 3 chunks in A3C, we observed less time spent blocked than updating.} 

\section{Experiments}
\label{sec:experiments}

We used the Atari-2600 domain to study the scaling characteristics of highly parallelized RL, investigating the following: 1) How efficient is synchronized sampling, and what speeds can it achieve?  2) Can policy gradient and Q-learning algorithms be adapted to learn using many parallel simulator instances without diminishing learning performance? 3) Can large-batch training and/or asynchronous methods speed up optimization without worsening sample complexity?

In all learning experiments, we maintained the original \textit{training intensity}--meaning average number of training uses of each sampled data point.  For A3C, PPO, and DQN+variants, the reference training intensities are 1, 4, and 8, respectively.

All learning curves shown here are averages over at least two random seeds.  For policy gradient methods, we tracked online scores, averaging over the most recent 100 completed trajectories.  For DQN and variants, we paused every 1-million steps to evaluate for up to 125,000 steps, with maximum path length of 27,000 steps, as is standard.  The appendices contain learning curves and experiment details beyond those we highlight here, including additional hyperparameter adjustments.

\subsection{Sampling}
\begin{figure}[ht]
    \begin{center}
    \subfigure[]{
        \label{fig:sampler_diagram}
        \includegraphics[width=0.45\textwidth]{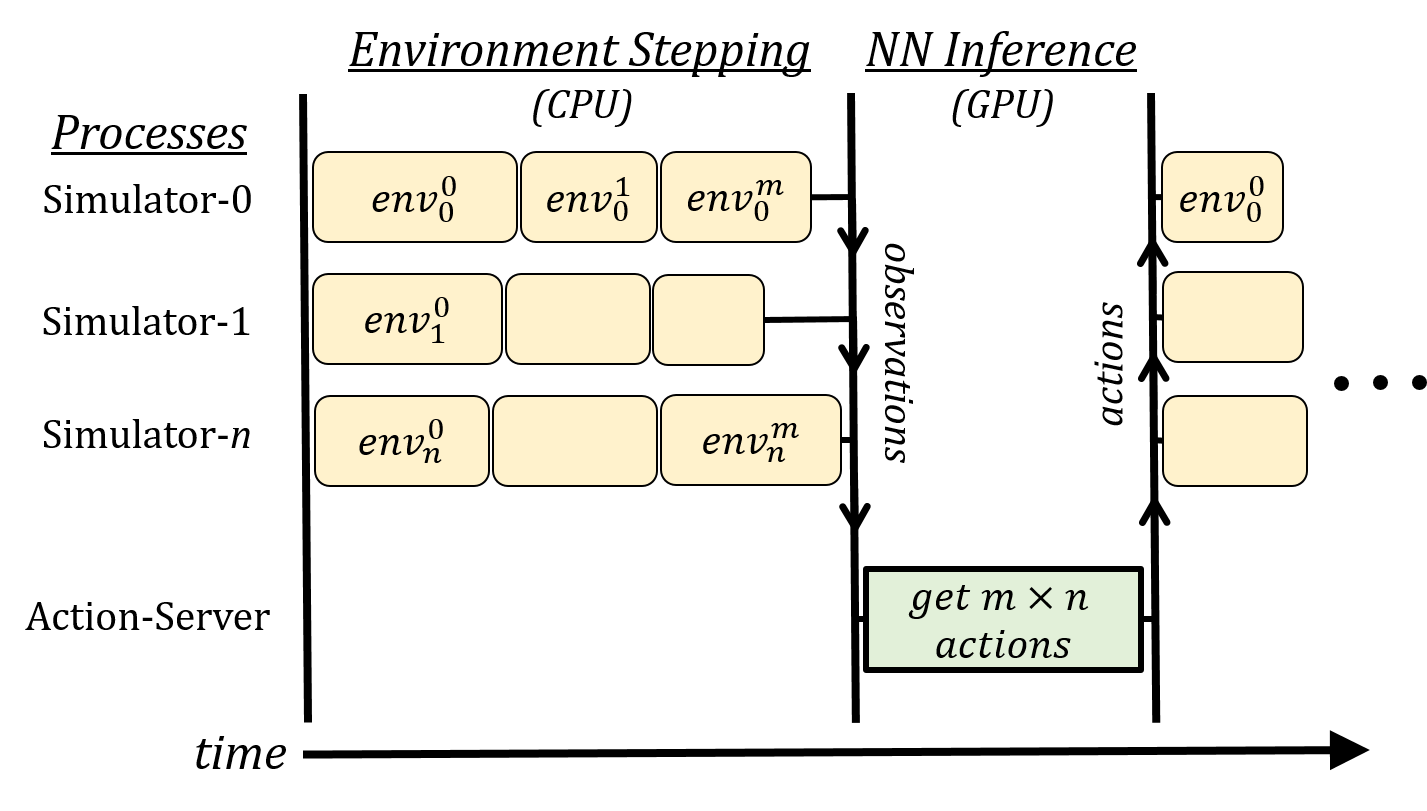}}
        \qquad
    \subfigure[]{
        \label{fig:sampler_speed}
        \includegraphics[width=0.375\textwidth]{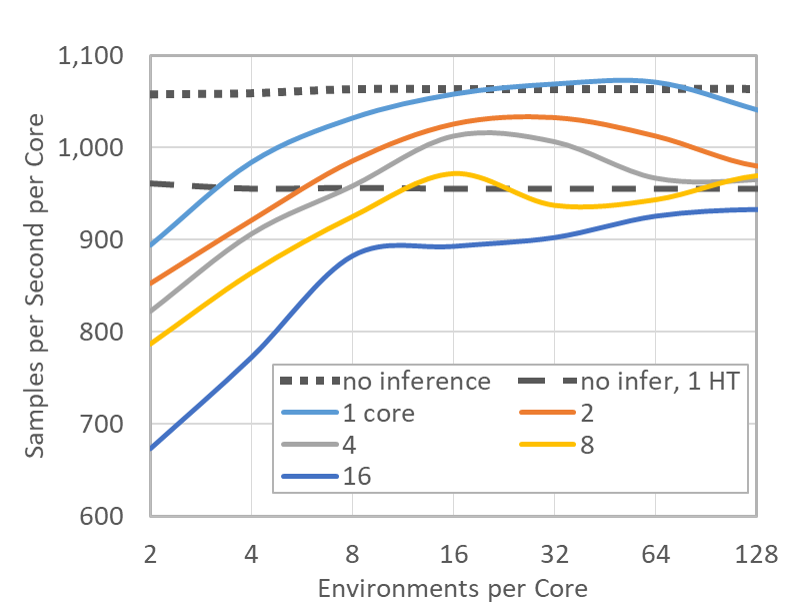}}
    \caption{Synchronized sampling (a) schematic: $n$ parallel simulation processes, each with $m$ sequential simulator instances, interacting synchronously with GPU-based action-server process (alternation not shown) (b) speed vs number of simulators per core, using 1 GPU.  Running multiple simulators per core mitigates synchronization losses and hides NN inference time, resulting in higher throughput.}
    \end{center}
\end{figure}

A series of sampling-only measurements demonstrated that despite the potential for stragglers, the synchronized sampling scheme can achieve good hardware utilization.  First, we studied the capacity of a single GPU at serving inferences for multiple environments.  Figure~\ref{fig:sampler_speed} shows measurements running a trained A3C-Net policy on a P100 GPU while playing \textsc{Breakout}.  Aggregate sampling speed, normalized by CPU core count, is plotted as a function of the number of (sequential) Atari simulators running on each  core.\footnote{Intel Turboboost was disabled for this test only, keeping the clock speed of every core at 2.2 GHz.}  The minimum was 2 simulators per core by the alternating scheme.  Different curves represent different numbers of CPU cores running simulations.  For reference, we include the sampling speed of a single core running without inference--the dashed line for a single process, and the dotted line one process on each of the two Hyperthreads.  Running with inferences and with a single core, the sampling speed increased with simulator count until the inference time was completely hidden.  Synchronization losses appeared for higher core count.  But at as little as 8 environments per core, the GPU supported even 16 CPU cores running at roughly 80\% of the inference-free speed.  

Next, we measured the sampling-only speed of the same A3C-Net playing \textsc{Breakout} parallelized across an entire 8-GPU, 40-core server.  At simulator counts of 256 (8 per core) and above, the server achieved greater than 35,000 samples per second, or 500 million emulator frames per hour, confirming scalability.  The appendix contains a table of results for other simulator counts.

\subsection{Learning with Many Simulator Instances}
To leverage the high throughput of parallel sampling, we investigated ways to adapt existing deep RL algorithms to learn with many simulator instances.  The following findings show that only minor changes suffice to adapt all algorithms and maintain performance.  We experimented with different techniques for each algorithm, which we describe and evaluate here.  Interestingly, scaling affects synchronous and asynchronous learning somewhat differently.

\textbf{Starting State Decorrelation:} Learning failed very early in some policy gradient experiments with many simulators.  We found correlation in starting game states to result in large but poorly informed learning signals, destabilizing early learning.  We correct this by stepping every simulator through a random number of uniform-random actions during experiment initialization.  When taking this measure, we found learning rate warmup \cite{Imagenet_1_Hour} to have no further effect.  While training, game resets proceed as usual. 

\begin{figure*}[ht]
    \begin{center}
    \includegraphics[width=0.9\textwidth]{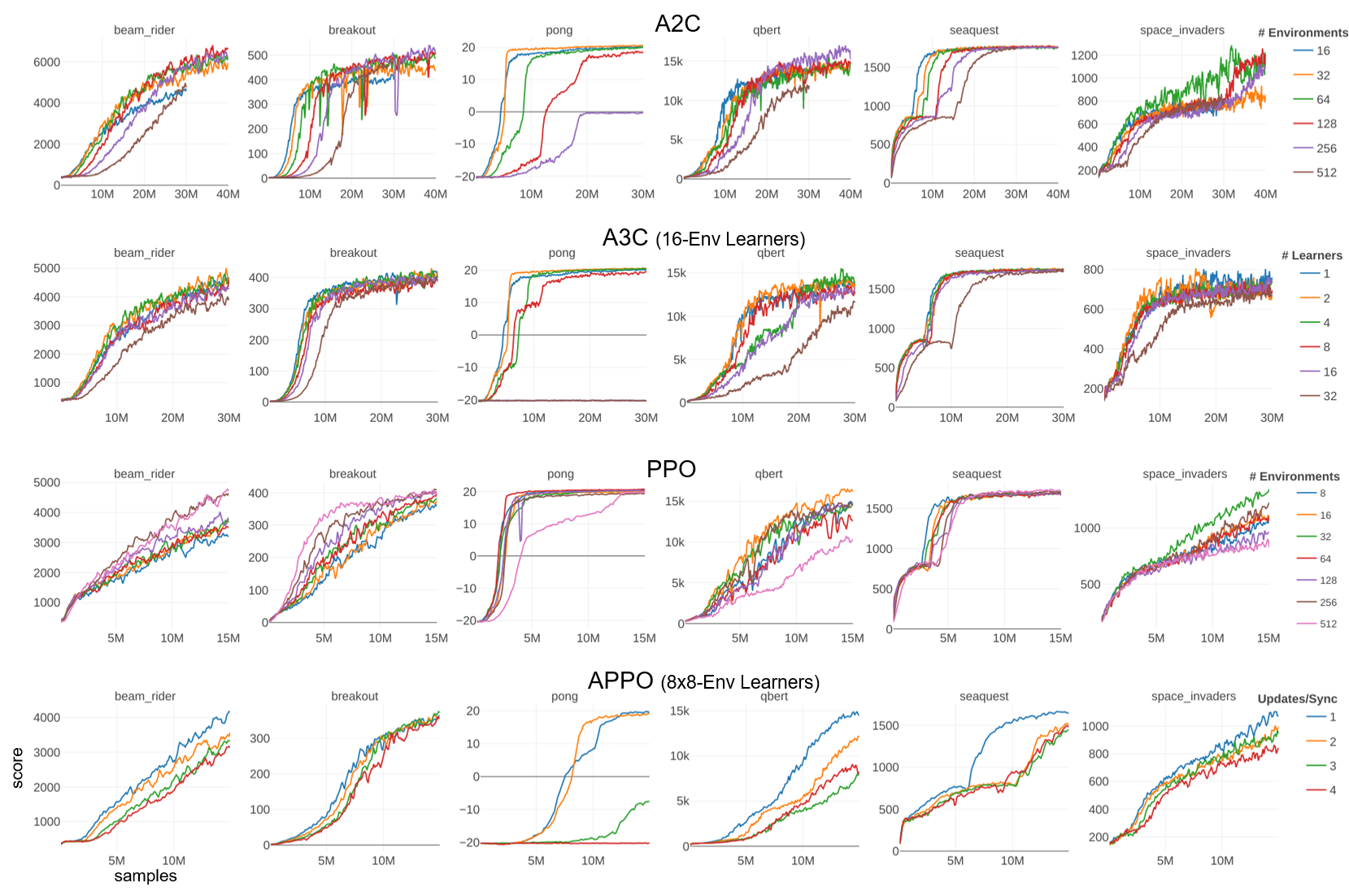}
    \caption{Scaling investigations for policy gradient algorithms: game scores vs aggregate sample count.  Top) A2C with various batch sizes (proportional to environment count), Upper) A3C with various numbers of 16-environment learner processes, Lower) PPO with varied number of simulators, Bottom) Asynchronous PPO, 8 learners with varied communication period.  In most cases, the scaled/adapted versions match the baseline performance.}
    \label{fig:pg_scaling}
    \end{center}
\end{figure*}

\textbf{A2C:}  The optimization batch size grows with the number of simulators (keeping the sampling horizon fixed).  Correspondingly fewer parameter update steps are made per sample gathered.  Unlike in \cite{ClementeMC17}, we found that increasing the learning rate with the \textit{square root} of the batch size worked best across a test set of games.  The top panel of Figure~\ref{fig:pg_scaling} shows learning curves vs total sample count, with simulator count ranging from 16 to 512 (batch size 80 to 2,560).  Game scores were largely unchanged, although a gradual decay in sample efficiency remained for large simulator counts.

\textbf{A3C:}  An asynchronous adaptation we tested used a 16-environment A2C agent as the base sampler-learner unit.  Figure~\ref{fig:pg_scaling} shows learning curves vs aggregate sample count for numbers of learners ranging from 1 to 32,\footnote{Learner counts in excess of 8 were run with multiple separate learners sharing GPUs.} corresponding to 16 to 512 total simulators.  The resulting learning curves were nearly indistinguishable in most cases, although some degraded at the largest scales.

\textbf{PPO:}  The large batch size already used to benchmark PPO (8-simulator x 256-horizon = 2,048) provides a different route to learning with many simulators: we decreased the sampling horizon such that the total batch size remained fixed.  Figure~\ref{fig:pg_scaling} shows learning curves vs sample count for simulator counts ranging from 8 to 512, with corresponding sampling horizons from 256 down to 4 steps.  Successful learning continued to the largest scale. 

\textbf{APPO:}  We also experimented with an asynchronous version of PPO, using an 8-simulator PPO agent as the base learner unit. The bottom panel in Figure~\ref{fig:pg_scaling} shows learning curves from a study of 8 learners running on 8 GPUs, with varying communication frequency.  Standard PPO uses 4 gradient updates per epoch, and 4 epochs per optimization; we experimented with 1-4 gradient updates between synchronizations (update rule provided in supplementary material).  We found it helpful to periodically pull new values from the central parameters during sampling, and did this with a horizon of 64 steps in all cases (thus decreasing policy lag inherent in asynchronous techniques, lag made acute by PPO's less frequent but more substantial updates).  In several games, the learning remained consistent, showing it is possible to reduce communication in some asynchronous settings.

\textbf{DQN + Variants:}  We organized the experience replay buffer by simulator.  The total buffer size remained at 1 million transitions, so a correspondingly shorter history was held for each simulator.  We observed learning performance to be largely independent of simulator count up to over 200, provided the number of update steps per optimization cycle is not too high (large batch size ameliorates this).

\subsection{Q-Value Learning with Large Training Batches}

\begin{figure*}[ht]
    \begin{center}
    \centerline{\includegraphics[width=\textwidth]{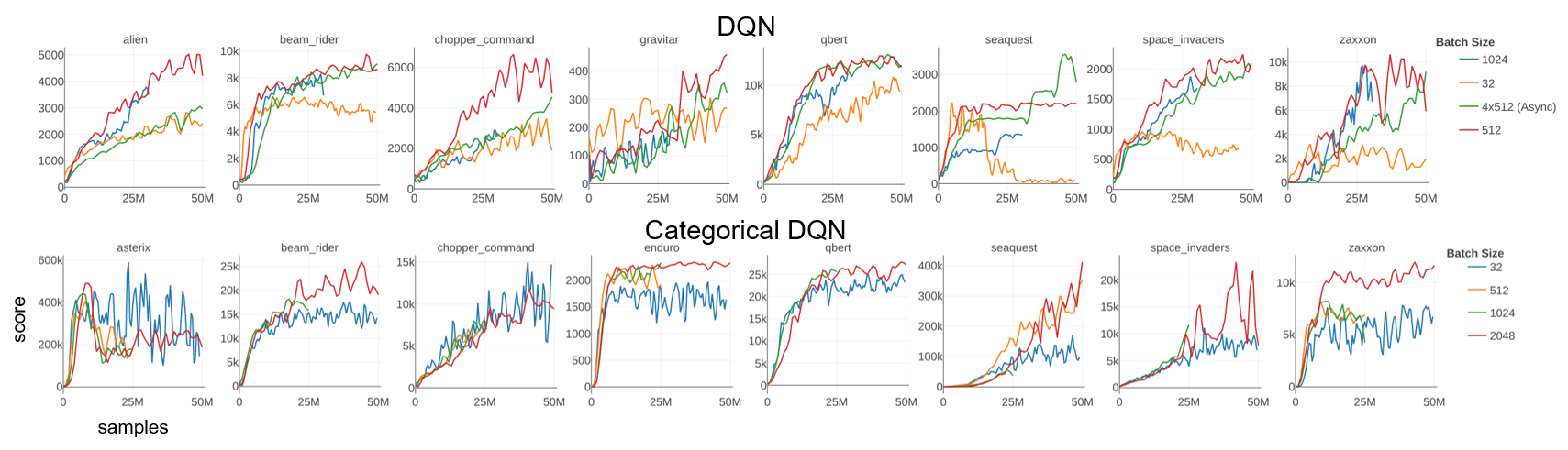}}
    \caption{Scaling investigations for DQN (top) and Categorical-DQN (bottom): game scores vs sample count. Both learn well using training batch sizes as large as 512; Categorical-DQN succeeds using up to 2,048.}
    \label{fig:dqn_scaling}
    \end{center}
    \vskip -0.2in
\end{figure*}

\textbf{DQN:}  We experimented with batch sizes ranging from the standard 32 up to 2,048.  We found consistent learning performance up to 512, beyond which, it became difficult to find a single (scaled) learning rate which performed well in all test games. In several games, a larger batch size improved learning, as shown in Figure~\ref{fig:dqn_scaling}.  We also found asynchronous DQN to learn well using up to 4 GPU learners, each using batch size 512. 

\textbf{Categorical DQN:}  We found Categorical DQN to scale further than DQN.  The lower panel of Figure~\ref{fig:dqn_scaling} shows learning curves for batch sizes up to 2,048, with no reduction in maximum scores.  This was possibly due to richer content of the gradient signal.  Notably, learning was delayed for the largest batch sizes in the game \textsc{Seaquest}, but greater maximum scores were eventually reached.  Due to use of the Adam optimizer, scaling of the learning rate was not necessary, and we return to study this surprising result shortly.

\textbf{$\epsilon$-Rainbow:}  Despite its use of distributional learning, $\epsilon$-Rainbow lost performance above batch size 512 in some games.  Scores at this batch size roughly match those reported in the literature for batch size 32 \cite{Hessel2017rainbow} (curves shown in appendix).   

\begin{table}[t]
    \caption{Average and median human-normalized scores across 49 games, including baseline and scaled configurations.  Scaled versions tend to match un-scaled versions. PG: 25M steps, DQN: 50M steps.}
    \label{tbl:normalized_scores}
    \centering
    \begin{tabular}{lcc}
        \toprule
        Algorithm-Scale & Average & Median \\
        \toprule
        A2C 16-env & 2.5 & 0.65 \\
        A2C 128-env & 8.0 & 0.48 \\
        A3C 3$\times$16-env & 5.4 & 0.53 \\
        \midrule
        PPO 8-env & 2.8 & \bf{1.2} \\ 
        PPO 128-env & \bf{9.8} & 0.97 \\ 
        APPO 8$\times$8-env & 8.8 & \bf{1.2} \\ 
        \toprule
        DQN-32\textsuperscript{*} & 2.4 & 0.96 \\
        DQN-512 & 9.8 & 1.4 \\
        \midrule
        CatDQN-32\textsuperscript{**} & \bf{25.5} & 2.1 \\
        CatDQN-2048 & 22.1 & \bf{2.9} \\
        \midrule
        $\epsilon$-Rainbow-512 & 22.4 & 2.5 \\ 
        \bottomrule
        \multicolumn{3}{l}{\textsuperscript{*}\footnotesize{\textup{from \cite{van2016double}}}} \\ \multicolumn{3}{l}{\textsuperscript{**}\footnotesize{\textup{from \cite{Bellemare2017distributional}}}}
    \end{tabular}
\end{table}

\subsection{Learning Speed}
\begin{figure*}[ht]
    \begin{center}
    \centerline{\includegraphics[width=\textwidth]{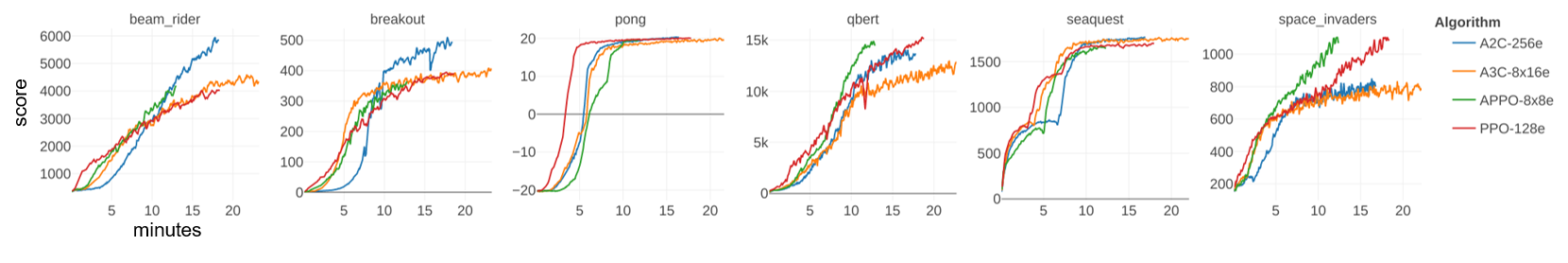}}
    \vspace{-0.2in}
    \caption{Policy gradient algorithms using an entire 8-GPU, 40-core server for a single learning run: game scores vs time, in minutes.  Asynchronous and synchronous versions learn successful game strategies in under 10 minutes.}
    \label{fig:pg_speed}
    \end{center}
    \vspace{-0.2in}
\end{figure*}

We investigated the learning speeds obtainable when running an 8-GPU, 40-core server (P100 DGX-1) to learn a single game, as an example large-scale implementation. Figure~\ref{fig:pg_speed} shows results for well-performing configurations of the policy gradient methods A2C, A3C, PPO, and APPO.  Several games exhibit a steep initial learning phase; all algorithms completed that phase in under 10 \textit{minutes}.  Notably, PPO mastered Pong in 4 minutes.  A2C with 256 environments processed more than 25,000 samples per second, equating to over 90 million steps per hour (360 million frames).  Table~\ref{tbl:dgx_speed} lists scaling measurements, showing greater than 6x speedup using 8 GPUs relative to 1.

\begin{table}[t]
    \caption{Hours to complete 50 million steps (200M frames) by GPU and CPU count.  A2C/A3C used 16 environments per GPU, PPO/APPO used 8 (DQN batch sizes shown).}
    \label{tbl:dgx_speed}
    \hskip -0.14in
    \begin{small}
    \begin{sc}
    \begin{tabular}{lcccc}
        \toprule
         & \multicolumn{4}{c}{\textbf{\# GPU} (\# CPU)}  \\
        Algo & \textbf{1} (5) & \textbf{2} (10) & \textbf{4} (20) & \textbf{8} (40) \\
        \midrule
        A2C & 3.8 & 2.2 & 1.2 & 0.59 \\
        A3C & -- & 2.4 & 1.3 & 0.65 \\
        PPO & 4.4 & 2.6 & 1.5 & 1.1 \\
        APPO & -- & 2.8 & 1.5 & 0.71 \\
        \toprule
        Algo-B.S. &  \textbf{1} (5) & \textbf{2} (10) & \textbf{4} (20) & \textbf{8} (40) \\
        \midrule
        DQN-512 & 8.3 & 4.8 & 3.1/3.9\textsuperscript{*} & 2.6 \\
        $\epsilon$-Rnbw-512 & 14.1 & 8.6 & 6.6 & 6.4 \\
        CatDQN-2k & 10.7 & 6.0 & 2.8 & 1.8 \\
        \midrule
        \multicolumn{4}{l}{\textsuperscript{*}\footnotesize{\textup{Asynchronous}}}
    \end{tabular}
    \end{sc}
    \end{small}
    \vskip -0.05in
\end{table}

We ran synchronous versions of DQN and its variants, with training times shown in Table~\ref{tbl:dgx_speed}.  Using 1 GPU and 5 CPU cores, DQN and $\epsilon$-Rainbow completed 50 million steps (200 million frames) in 8 and 14 hours, respectively--a significant gain over the reference times of 10 days.  These learning speeds are comparable to those in \cite{horgan2018distributed}, which used 1 GPU and 376 CPU cores (see e.g. Figure 2 therein for 10-hour learning curves).  Using multiple GPUs and more cores sped up our implementations.  By virtue of a larger batch size, Categorical-DQN scaled best and completed training in under 2 hours using the entire server, a speedup of over 6x relative to 1 GPU.  DQN and $\epsilon$-Rainbow, however, experienced diminishing returns beyond 2 GPUs.  We were unable to find asynchronous configurations that further boosted learning speed without curbing performance in some games (we only tested fully-communicating algorithms).  Opportunities may exist to improve on our scaling.

\subsection{Effects of Batch Size on Optimization}

Possible factors limiting training batch sizes include: 1) reduced exploration, since fewer different networks are exercised in the environment, and 2) difficulties in numerical optimization of network weights.  We conducted experiments to begin to identify these factors.

\textbf{Secondary-Learner Experiment:}  We configured a secondary DQN learner to train using only the replay buffer of a normal DQN agent.  The secondary learner was initialized with the same parameter values as the primary, ``sampler-learner'', and the two networks trained simultaneously, at the same rate of data consumption.  Each sampled its own training batches.  In the game of \textsc{Breakout}, 64- and 2048-sampler-learners achieved the same score, but the 2048-learner required more samples, despite using the fastest stable learning rate (the number refers to training batch size).  When training a 64-secondary-learner using a 2048-sampler-learner, the secondary learner's score tracked that of the primary.  In the reverse scenario, however, the 2048-secondary-learner failed to learn. We posit this was due to the slower optimization of the decreased number of parameter updates--it was unable to track the rapid changes to the Q-value estimates near initialization and became too off-policy to learn.  In the same test using two 256-learners, their scores matched.  Had the 2048-secondary-learner out-paced the 2048-sampler-learner, it would have suggested exploration to be a more important factor than optimization.  See the supplementary materials for figures.

\textbf{Update Rule:}  We conducted an experiment to isolate the effect of update rule on optimization in Categorical DQN.  We found the Adam \cite{adam} formula to be superior to RMSProp \cite{rmsprop} in providing large-batch learners with capability to traverse parameter-space during learning.  When comparing agents achieving the same learning curves, those using smaller batch sizes (and hence performing more update steps) tended to have larger parameter vector-norms at all points in training.  Unlike RMSProp, the Adam rule resulted in a fairly tight spread in parameter norms between batch sizes without changing the learning rate.  This explains the lack of need to scale the learning rate in Categorical DQN and $\epsilon$-Rainbow, and indicates that the update rule plays an important role in scaling.  Further details, including trends in convolutional and fully connected layers, appear in an appendix.

\textbf{Gradient Estimate Saturation:}  Using A2C, we measured the relation between the normal, full-batch gradients and gradients computed using only half of the batch, at each iteration.  For small-batch agents, the average cosine-similarity between the full- and half-batch gradients measured near $1/\sqrt{2}$.  This implies the two half-batch gradients were orthogonal, as are zero-centered random vectors in high-dimensional spaces.  For large-batch learners (e.g. 256 environments), however, the cosine similarity increased significantly above $1/\sqrt{2}$.  Saturation of the gradient estimate was clearly connected to worsened sample efficiency as in the learning curves in the top panel of Figure~\ref{fig:pg_scaling}.

\section{Conclusions and Discussion}
\label{sec:conclusion}

We have introduced a unified framework for parallelizing deep RL that uses hardware accelerators to achieve fast learning.  The framework is applicable to a range of algorithms, including policy-gradient and Q-value learning methods.  Our experiments show that several leading algorithms can learn a variety of Atari games in highly parallel fashion, without loss of sample complexity and at unprecedented wall-clock times.  This result indicates a promising direction to significantly boost experiment scale.  We will release the code-base.

We note several directions for extension of this framework.  First is to apply it to domains other than Atari, especially ones involving perception.  Second, our framework is likely to scale favorably to more sophisticated neural network agents, due to GPU acceleration of both inference and training.  Moreover, as network complexity increases, scaling could become easier, as GPUs may run efficiently with smaller batch sizes, although communication overhead could worsen.  Reduced-precision arithmetic could hasten learning--a topic yet to be explored in deep RL due to use of CPU-based inference.  The current, single-node implementation may be a building block for distributed algorithms.

Questions remain as to the extent of parallelization possible in deep RL.  We have not conclusively identified the limiting factor to scaling, nor if it is the same in every game and algorithm.  Although we have seen optimization effects in large-batch learning, other factors remain possible.  Limits to asynchronous scaling remain unexplored; we did not definitively determine the best configurations of these algorithms, but only presented some successful versions.  Better understanding may enable further gains in scaling, which is a promising direction for the advancement of deep RL.   

\section*{Acknowledgements}
Adam Stooke gratefully acknowledges the support of the Fannie \& John Hertz Foundation.  The DGX-1 used for this research was donated by the NVIDIA Corporation.  We thank Fr\'{e}d\'{e}ric Bastien and the Theano development team \cite{Theano} for their framework and helpful discussions during development of GPU-related methods.  Thanks to Rocky Duan \textit{et al} for the \textit{rllab} code-base \cite{rllab} out of which this work evolved.

\bibliographystyle{sysml2019}
\bibliography{main.bbl}

\onecolumn
\appendix 
\centerline{\LARGE\bf Supplementary Materials}

\section{Experiment Details}

\subsection{Atari Frame Processing}
Our frame pre-processing closely resembles that originally described in the original DQN publication.  The sole difference is that we abandon the square frame dimensions in favor of simply downsizing by a factor of 2, which provides crisp images at minimal computational cost.  Before downsizing, we crop two rows, making the final image size $104\times80$.  This simplifies selection of convolution size, stride, and padding.  Otherwise, we keep all standard settings.  For Q-learning experiments, we used the standard 3-convolutional-layer network (DQN-Net) or its algorithm-specific variants, and for policy gradients the standard 2-convolutional-layer feed-forward network of (A3C-Net).  The second (and third) convolution layers have padding 1, so the convolution output is always $12\times9$.

\subsection{DGX-1 Sampling Speed}
Table~\ref{tbl:sampling_speed} shows results of DGX-1 sampling speed, playing \textsc{Breakout} using a trained A3C-Net, for various total simulator counts.  In the synchronized setting, a barrier was placed across all GPU processes every five steps (mimicking the optimization in A2C).  Otherwise, each GPU and its associated cores ran independently.

\begin{table}[ht]
\caption{Sampling speeds on the DGX-1 with A3C-Net, by total simulator count, in thousands of samples per second.}
\label{tbl:sampling_speed}
\vskip 0.1in
\begin{center}
\begin{small}
\begin{sc}
\begin{tabular}{lcc}
\toprule
\# Sims (per Core) & Sync & Async \\
\midrule
64 (2)    & 29.6 & 31.9 \\
128 (4) & 33.0 & 34.7 \\
256 (8)    & 35.7 & 36.7 \\
512 (16)    & 35.8 & 38.4  \\
\bottomrule
\end{tabular}
\end{sc}
\end{small}
\end{center}
\vskip -0.2in
\end{table}

\subsection{Hyperparameters for Scaling}
\textbf{A2C:}  We used RMSProp with a base learning rate of $7\times10^{-4}$ for 16 environments (e.g., scaled up to $3\times10^{-3}$ for 512 environments).

\textbf{A3C:} We found no hyperparameter adjustments to be necessary.

\textbf{PPO:}  We did not change any optimization settings.

\textbf{APPO:}  Each 8-simulator asynchronous agent used the same sampling horizon (256) and update sequence as original PPO.  Relative to PPO, we introduced gradient-norm-clipping, reduced the learning rate by a factor of four, and removed the learning rate schedule, all of which benefited learning.

\textbf{DQN:}  When growing the simulator count and batch size in DQN, we maintained training intensity by adjusting the sampling horizon and the number of update steps per optimization phase.  For the batch sizes 32, 512, and 1024, we used learning rates $2.5, 7.5, 15\times10^{-4}$, respectively.  Hyperparameters other than learning rate were as in original publication.

\textbf{Categorical DQN:}  We used a learning rate of $4.2\times10^{-4}$ at all batch sizes 256 and above.  We employed the published setting for epsilon in the Adam optimizer: $0.01/L$, where $L$ is the batch size.

\textbf{$\epsilon$-Rainbow:} We used the published hyperparameters without scaling the learning rate.  We scaled epsilon in the Adam update rule as $0.005/L$, with $L$ the batch size.

\newpage
\section{Update Rule for Multi-Step Asynchronous Adam}
Our asynchronous PPO experiments used the update rule described here, which permits multiple local gradient steps per synchronization with the central parameters.  The usual Adam update rule \cite{adam} is the following.  It has fixed hyperparameters $r$, $\beta_1$, $\beta_2$, and $\epsilon$; $g$ stands for the gradient; and all other values except the network parameters $\theta$ are initialized at $0$:
\begin{align*}
t &\leftarrow t + 1 \\
a &\leftarrow r\frac{\sqrt{1-\beta_2^t}}{1-\beta_1^t} \\
m &\leftarrow \beta_1m + (1-\beta_1)g \\
v &\leftarrow \beta_2v + (1 - \beta_2)g^2 \\
s &\leftarrow \frac{am}{\sqrt{v}\epsilon} \\
\theta &\leftarrow \theta - s \quad . \\
\end{align*}

We kept these rules for making local updates and introduced the additional, local accumulation variables, also zero-initialized:
\begin{align*}
a_g &\leftarrow \beta_1a_g + g \\
a_{g^2} &\leftarrow \beta_2a_{g^2} + g^2 \\
a_s &\leftarrow a_s + s \quad .
\end{align*}
When applying an update to the central parameters, denoted with a tilde, we used the following assignments:
\begin{align*}
\theta, \tilde{\theta} &\leftarrow \tilde{\theta} - a_s \\ 
m, \tilde{m} &\leftarrow \beta_1^n\tilde{m} + (1-\beta_1)a_g \\ 
v, \tilde{v} &\leftarrow \beta_2^n\tilde{v} + (1-\beta_2)a_{g^2} \\ 
a_g, a_{g^2}, a_s &\leftarrow 0
\end{align*}

where $n$ is the number of local gradient steps taken between synchronizations. This rule reduces to the usual Adam update rule in the case of a single learner thread.

\newpage
\section{Figures for Secondary-Learner Experiment (DQN)}

\begin{figure}[h]
\begin{center}
\subfigure[]{
\label{fig:twobs_2048_64}
\includegraphics[width=0.35\textwidth]{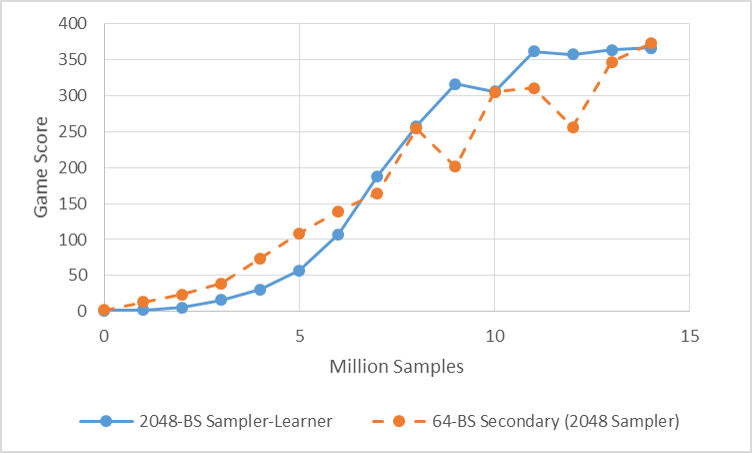}}
\qquad
\subfigure[]{
\label{fig:twobs_64_2048}
\includegraphics[width=0.35\textwidth]{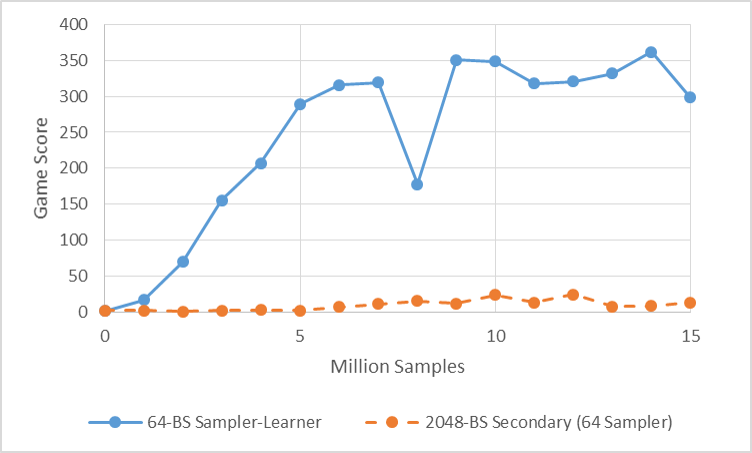}}
\caption{Learning the game \textsc{Breakout} with a secondary-learner using only the replay buffer of the normal, sampler-learner, both using DQN.  a) The 64-batch-size secondary-learner kept pace with its 2048-batch-size sampler-learner, but b) the 2048-batch-size secondary learner failed to track its 64-batch-size sampler-learner or even learn at all. (Curves averaged over two random trials.)}
\end{center}
\end{figure}

\begin{figure}[h]
\begin{center}
\subfigure[]{
\label{fig:twobs_2048_64_pnorm}
\includegraphics[width=0.35\textwidth]{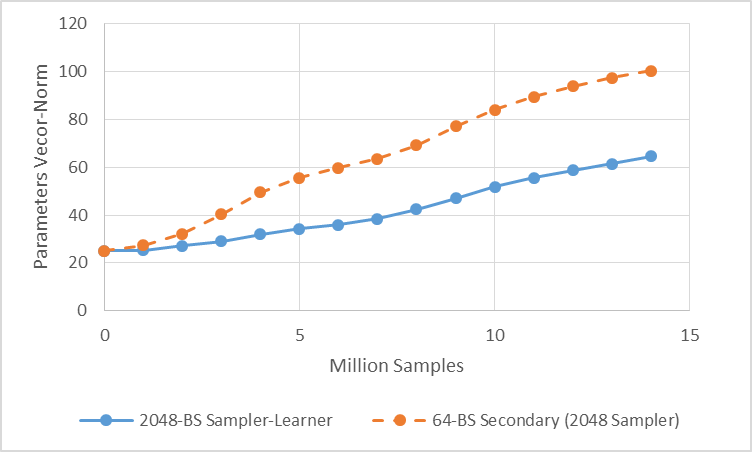}}
\qquad
\subfigure[]{
\label{fig:twobs_64_2048_pnorm}
\includegraphics[width=0.35\textwidth]{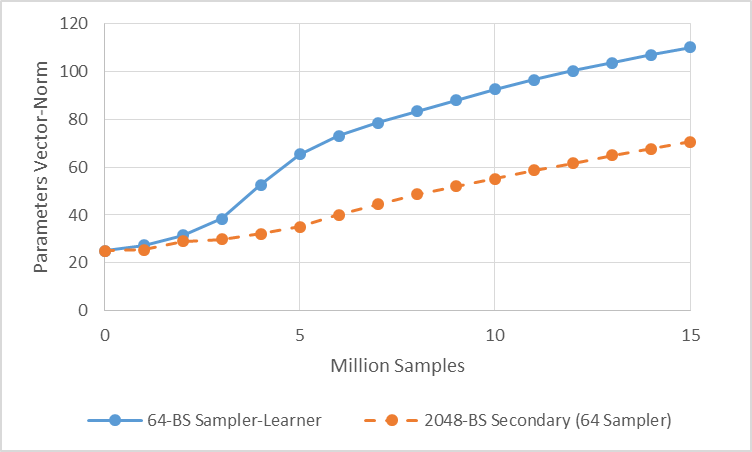}}
\caption{Neural network parameter vector-norms ($l$-2) during training.  In both cases, the large-batch learner lagged behind the small batch learner.  In b) the parameters of large-batch secondary-learner continued to grow while its game score remained nil.}
\end{center}
\end{figure}

\begin{figure}[H]
\begin{center}
\centerline{\includegraphics[width=0.3\textwidth]{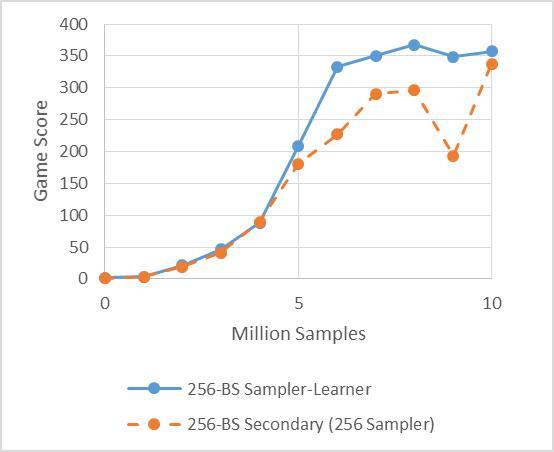}}
\caption{Learning the game \textsc{Breakout}, where a secondary-learner using the same batch-size as the sampler-learner tracked (albeit imperfectly) the game score, learning successfully. (Curves averaged over two random trials.)}
\label{fig:dqn_twobs_256_256}
\end{center}
\vskip -0.2in
\end{figure}

\newpage

\section{Observations on Update Rules and Batch Size Scaling}
We present observations of the effects of scaling the training batch size on neural net optimization under two different parameter update rules: Adam and RMSProp (RMSProp without momentum and only direct accumulation of the squared gradients, see e.g. \texttt{https://github.com/Lasagne/Lasagne/blob/master/lasagne/updates.py}).  We trained agents on the game \textsc{Q*Bert} with learning rates adjusted to yield very similar performance curves for all settings, and we tracked the $L$-2 vector-norms of several quantities during learning.  These included the gradients, parameter update steps, and the parameter values themselves.  As in all DQN experiments in this paper, the training intensity was fixed at 8, so that the number of parameter update steps during learning scaled inversely with the batch size.  Two random seeds were run for each setting.  

\begin{figure}[H]
\begin{center}
\centerline{\includegraphics[width=0.85\textwidth]{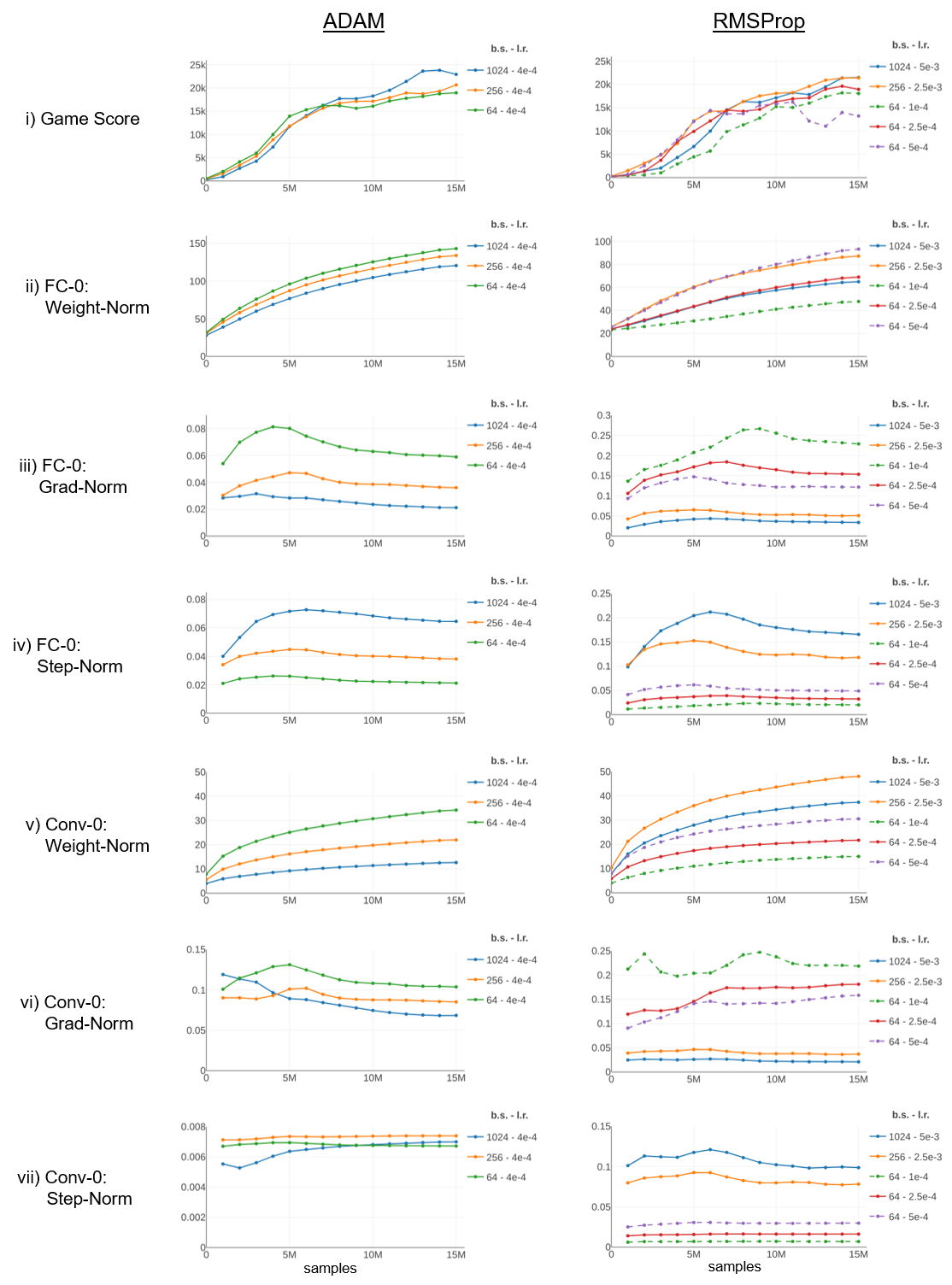}}
\caption{L-2 vector-norms of the parameters, gradients (average), and parameter steps (average) for the first convolution layer and the first fully connected layer while learning to play \textsc{Q*Bert} with Categorical DQN at various batch sizes: Adam vs RMSProp.}
\label{fig:catdqn_adam_vs_rmsprop}
\end{center}
\vskip -0.2in
\end{figure}

Although the game scores roughly matched throughout training, the exact solutions found at any point did not, as evidenced by the differing parameter norms.  No regularization was used.  The following paragraphs follow the panels in Figure~\ref{fig:catdqn_adam_vs_rmsprop}, where curves are labeled by batch-size and learning rate.  When viewing the network as a whole (i.e. norms of all weights and biases as a single vector), trends reflected those seen in FC-0, where most of the weights are.

\textbf{i) Learning Curves:} We controlled for game score, adjusting the learning rate as needed.  For the case of RMSProp with batch-size 64, we included a learning rate that was slightly too low ($1\times10^{-4}$), yielding slow learning and lower final score, and a learning rate that was slightly too high ($5\times10^{-4})$, yielding lower final score due to instability--these are the dashed lines in all panels.

\textbf{ii) Fully-Connected-0 Weights-Norm:}  
The Adam optimizer yielded fairly tight grouping despite using the same learning rate for all settings.  The RMSProp learner, on the other hand, needed to scale the learning rate by 20x between batch sizes 64 and 1,024, which then produced very similar norms.  At batch-size 64, slow / unstable learning was characterized by small / large norms, respectively.  The large norm of the batch-size 256 runs suggests this learning rate was likely near the upper limit of stability.   

\textbf{iii) Fully-Connected-0 Gradients-Norm:} Under both update rules, large batch sizes always produced smaller gradient vectors--reduced variance led to reduced magnitudes.  We also observed this pattern in policy gradient methods, when looking at the total gradient norm.  Here, the magnitude of the gradients depended inversely on the parameter norm; see the RMSProp 64-batch-size curves.  This effect was opposed and outweighed by the effect of batch size.

\textbf{iv) Fully-Connected-0 Step-Norm:}  The Adam optimizer yielded significantly bigger step sizes for the bigger batch learners, despite the smaller gradients.  RMSProp required an adjusted learning rate to produce the same effect.  Under both update rules, the amount of step size increase did not fully compensate for the reduction in step count, indicating that the larger batch learners followed straighter trajectories through parameter space.  RMSProp led to significantly larger steps overall, but despite this ended learning at smaller weights--its learning trajectories were apparently less direct, more meandering.  

\textbf{v) Convolution-0 Weights-Norm:} The Adam optimizer gave much greater spread in norms here than in the FC-0 layer; as batch size increased, the learning emphasis shifted away from Conv-0.  But in RMSProp the increased learning rate led the first convolution layer to grow larger for larger batch sizes, placing more emphasis on this layer. 

\textbf{vi) Convolution-0 Gradients-Norm:}  The Adam update rule produced an intriguing cross-over in gradient norm; the large batch learner actually started higher, bucking the trend seen in other cases.  The pattern under RMSProp matched that for FC-0.

\textbf{vii) Convolution-0 Step-Norm:}  Unlike for FC-0, the step norm did not change significantly with batch size under Adam.  RMSProp yielded a similar pattern as in FC-0.

Overall, the Adam optimizer appeared to compensate for batch size in the FC-0 layer, but less so in the Conv-0 layer, leading to de-emphasized learning in Conv-0 for large batches.  The increased learning rate in RMSProp compensated for batch size in the FC-0 layer and \textit{increased} the emphasis on learning in Conv-0.  This sort of pattern could have implications for learning representations vs game strategies.  Further study of these clear trends could yield insights into the causes of learning degradation and possible solutions for large batch RL.

\hfill
\newpage

\section{Additional Learning Curves}

\begin{figure}[H]
\begin{center}
\centerline{\includegraphics[width=1.05\textwidth]{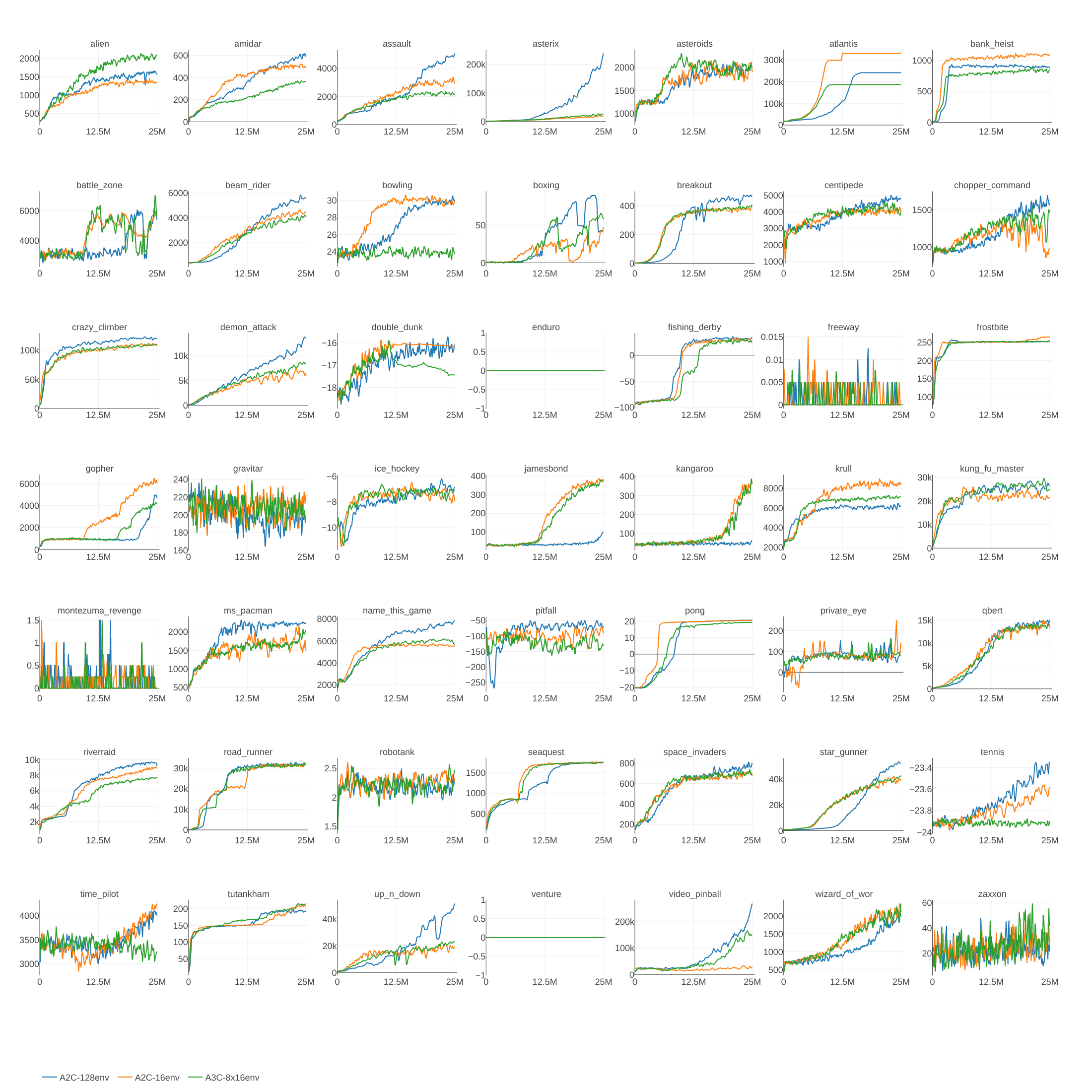}}
\caption{Learning curves for Advantage Actor-Critic: baseline (A2C-16env) and scaled configurations, including synchronous and asynchronous (to 25M steps = 100M frames).  Only in \textsc{Atlantis}, \textsc{Gopher}, and possibly \textsc{Krull} does the baseline stand out above both scaled versions.}
\label{fig:a2c_a3c_all_games}
\end{center}
\vskip -0.2in
\end{figure} 

\newpage

\begin{figure}[H]
\begin{center}
\centerline{\includegraphics[width=1.05\textwidth]{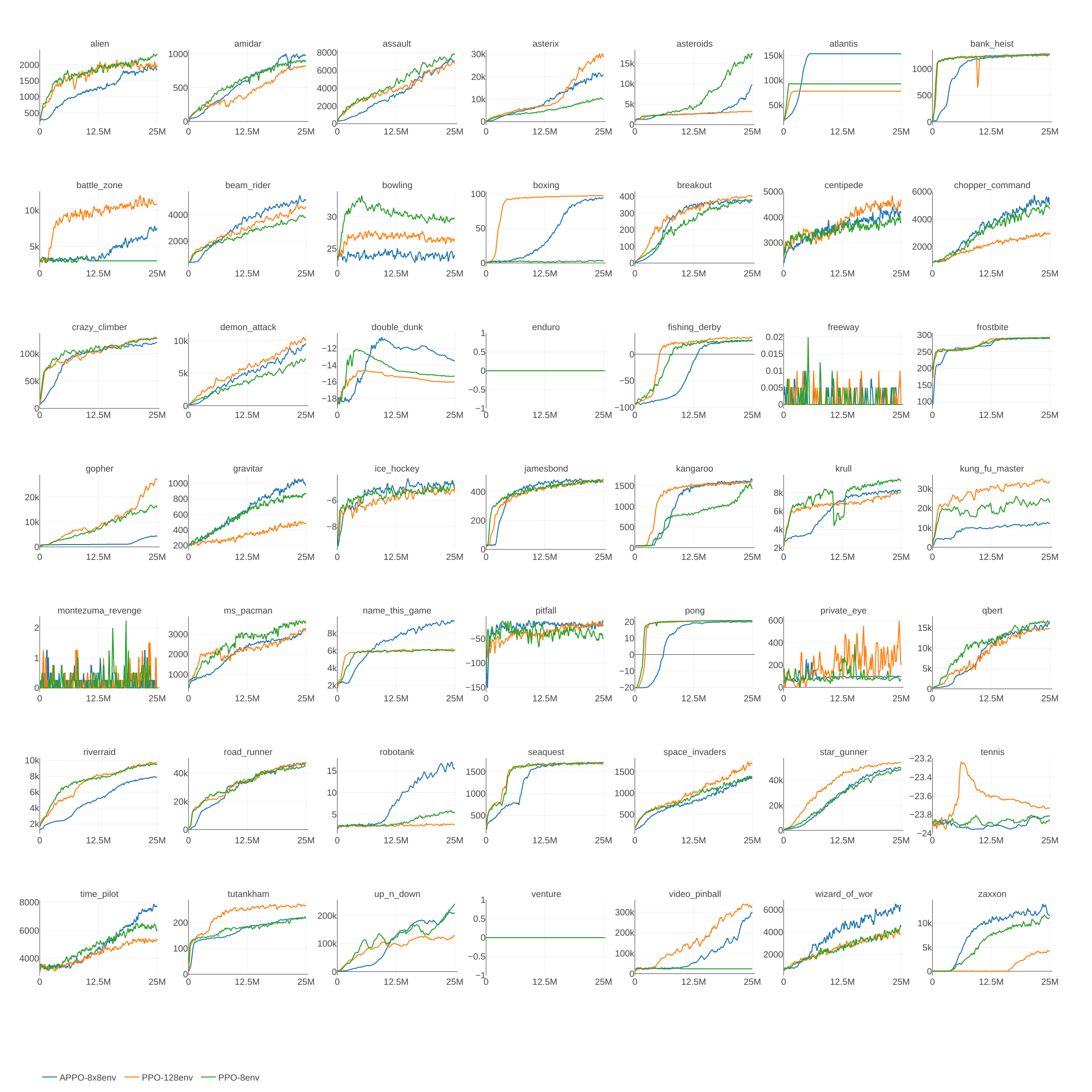}}
\caption{Learning curves for Proximal Policy Optimization: baseline (PPO-8env) and scaled configurations, including synchronous and asynchronous (to 25M steps = 100M frames).  Only in \textsc{Asteroids} and \textsc{Bowling} does the baseline stand out above both scaled versions.}
\label{fig:ppo_all_games}
\end{center}
\vskip -0.2in
\end{figure} 

\newpage

\begin{figure*}[ht]
\begin{center}
\centerline{\includegraphics[width=1.05\textwidth]{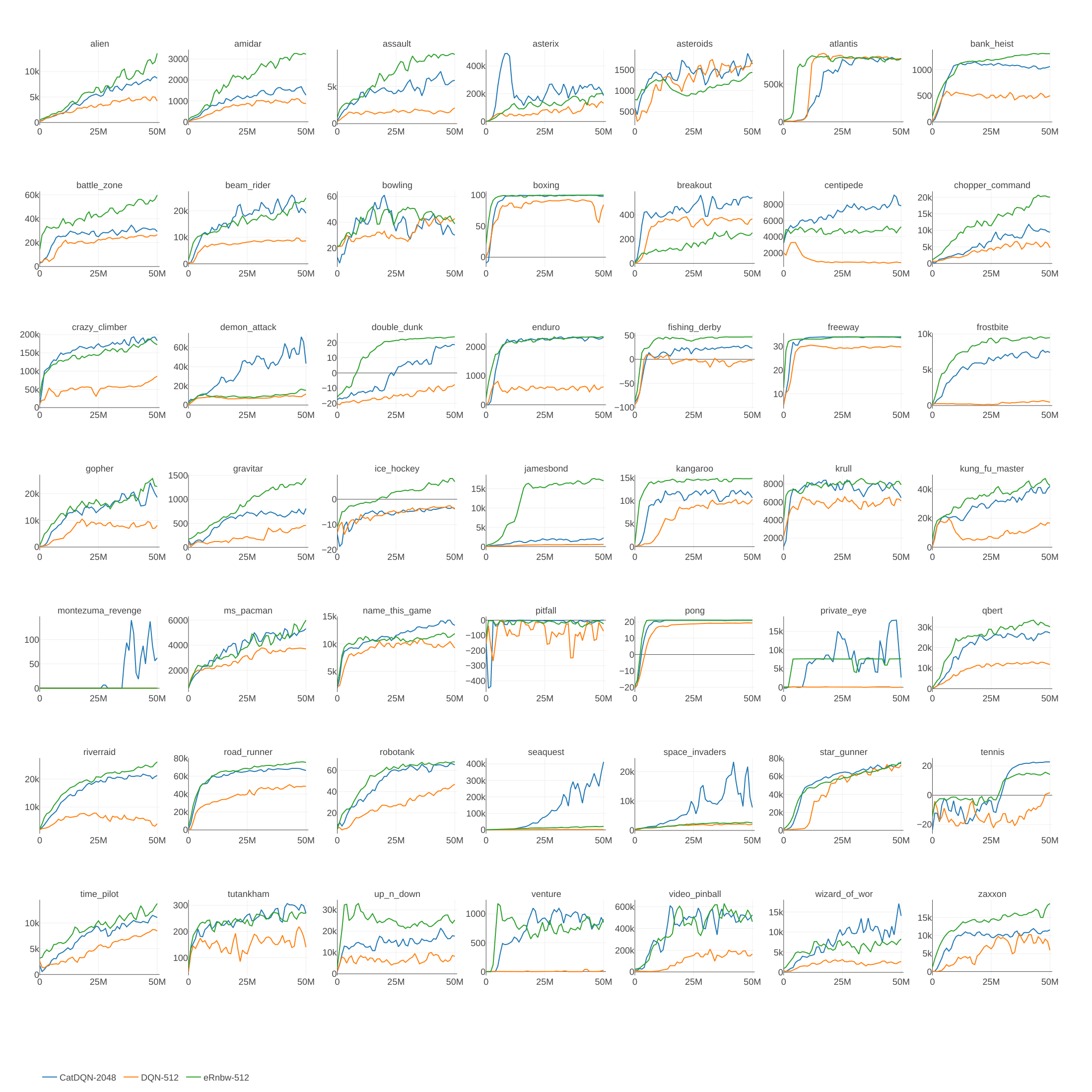}}
\caption{Learning curves for scaled versions of DQN (synchronous only): DQN-512, Categorical-DQN-2048, and $\epsilon$-Rainbow-512, where the number refers to training batch size (to 50M steps = 200M frames).  The anomalously low scores for $\epsilon$-Rainbow in \textsc{Breakout} also appeared for smaller batch sizes, but was remedied when setting the reward horizon to 1 or with asynchronous optimization (cause unknown; reward horizon 3 usually helped).}
\label{fig:dqns_all_games}
\end{center}
\vskip -0.2in
\end{figure*} 

\end{document}